\documentclass[sigconf]{acmart}
\usepackage{booktabs} 
\usepackage{todonotes}
\usepackage{url}
\usepackage{subfigure}

\setcopyright{rightsretained}

\renewcommand\footnotetextcopyrightpermission[1]{} 
\copyrightyear{2018} 
\acmYear{2018} 
\setcopyright{acmcopyright}
\acmConference[ICDSC '18]{International Conference on Distributed Smart Cameras}{September 3--4, 2018}{Eindhoven, Netherlands}
\acmPrice{15.00}
\acmDOI{10.1145/3243394.3243692}
\acmISBN{978-1-4503-6511-6/18/09}

\begin{document}

\author{George Plastiras}
\affiliation{%
  \institution{Department of Electrical and Computer Engineering,\\KIOS Research and Innovation Center of Excellence,\\University of Cyprus}
  \streetaddress{P.O. Box 1212}
  \city{Nicosia, Cyprus}
}
\email{gplast01@ucy.ac.cy}

\author{Christos Kyrkou}
\orcid{0000-0002-7926-7642}
\affiliation{%
  \institution{KIOS Research and Innovation Center of Excellence,\\University of Cyprus}
  \streetaddress{P.O. Box 1212}
  \city{Nicosia, Cyprus}
}
\email{kyrkou.christos@ucy.ac.cy}

\author{Theocharis Theocharides}
\affiliation{%
  \institution{Department of Electrical and Computer Engineering,\\KIOS Research and Innovation Center of Excellence,\\University of Cyprus}
  \streetaddress{1 Th{\o}rv{\"a}ld Circle}
  \city{Nicosia, Cyprus}
}
\email{ttheocharides@ucy.ac.cy}

\title[Efficient ConvNet-based Object Detection for UAVs by STP]{Efficient ConvNet-based Object Detection
for Unmanned Aerial Vehicles by Selective Tile Processing}

\begin{abstract}
Many applications utilizing Unmanned Aerial Vehicles (UAVs) require the use of computer vision algorithms to analyze the information captured from their on-board camera. Recent advances in deep learning have made it possible to use single-shot Convolutional Neural Network (CNN) detection algorithms that process the input image to detect various objects of interest. To keep the computational demands low these neural networks typically operate on small image sizes which, however, makes it difficult to detect small objects. This is further emphasized when considering UAVs equipped with cameras where due to the viewing range, objects tend to appear relatively small. This paper therefore, explores the trade-offs involved when maintaining the resolution of the objects of interest by extracting smaller patches (tiles) from the larger input image and processing them using a neural network. Specifically, we introduce an attention mechanism to focus on detecting objects only in some of the tiles and a memory mechanism to keep track of information for tiles that are not processed. Through the analysis of different methods and experiments we show that by carefully selecting which tiles to process we can considerably improve the detection accuracy while maintaining comparable performance to CNNs that resize and process a single image which makes the proposed approach suitable for UAV applications.
\end{abstract}


\maketitle

\colorlet{hl}{green!10!orange!90!}

\section{Introduction}

Unmanned Aerial Vehicles (UAVs) are becoming a widely used mobile camera platform with a wide range of applications such as asset inspection \cite{RadovicORUAV2017}, intelligent transportation systems \cite{KyrkouVisionDirected2018}, and disaster monitoring \cite{Bejiga.Avalanche.2017}. The cameras on-board UAVs are a reach source of information that can be processed through computer vision algorithms in order to extract meaningful information. In particular, detecting various objects is a key step in many such applications that can greatly enhance remote sensing and situational awareness capabilities. However, object detection based on state-of-the-art Convolutional Neural Networks (CNNs) is extremely computationally demanding and typically requires high-end computing hardware. When considering lightweight and low-cost UAVs as well as power-imposed constraints the use of such hardware can be prohibitive. In addition, for many applications processing on the edge -- locally embedded near the sensor -- is preferred over cloud-based approaches due to privacy or latency concerns, and for operation in remote areas where there is no internet connection.

Recent research has shown that by either developing specialized low-power hardware for embedded applications \cite{DATELowPowerImage2018} or by carefully designing the CNN structure \cite{DroNet2018} it is possible to achieve real-time processing with relatively high accuracy. However, a key issue with regards to the use of existing CNNs is that they have a specific receptive field which can limit the input image size. This is particularly important in certain cases such as cameras on-board UAVs where he altitudes that UAVs fly/hover, and the resulting resolution of objects in the image, make the detection of small objects even more challenging. However, increasing the CNN receptive field makes the learning process difficult since the problem dimensionality is increased and at the same time the computational cost also increases exponentially \cite{Suleiman:2017:CNN:Energy:Exp}. To tackle this issue it is necessary to reduce the data that needs to be processed from the original input image.

To this end the solution proposed in this work is to break the input image into smaller images (tiles) that each can then be processed by the CNN without degrading the resolution. Through this approach it is possible to detect smaller objects such as pedestrians with increased accuracy and at the same time does not incur significant computational cost. The proposed methodology is comprised of two mechanisms to analyze larger images than what the CNN receptive field can accept. A \textit{memory mechanism} is introduced to track detections in image areas we are not processing and an \textit{attention mechanism} is used to select image regions (tiles) for processing. We evaluate different approaches and attention mechanisms in order to identify the one that provides the best trade-off between accuracy and inference speed for use in UAV smart camera applications. The evaluation of different methods indicates that the overall performance remains high ($~20$ frames per second) with considerable improvement in accuracy (up to $70\%$) compared to approaches that resize the input image prior to detection.

\begin{figure}[t]
\centering
\includegraphics[width=1\linewidth]{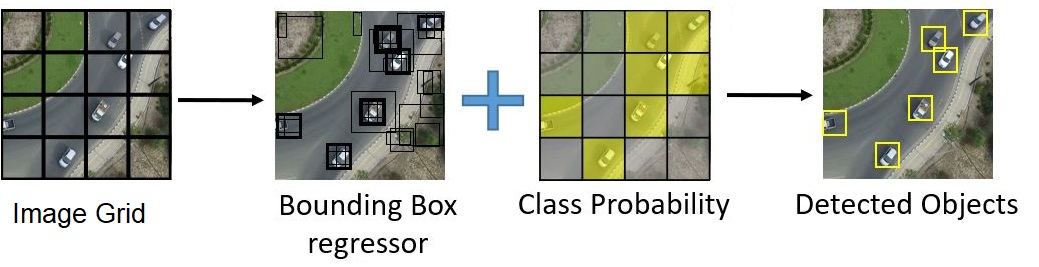}
\caption{Framework for single shot aerial object detection}
\label{fig:airYOLO}
\end{figure}

\section{Background and Related Work}
\label{sec:background}

\subsection{Single-Shot Object Detectors}
Object detection aims to find instances of objects from known classes in an image. This task typically yields the location and scale of the object in terms of a bounding box, together with a probability on its class. In the last decade, a lot of progress has been made on object detectors based on Convolutional Neural Networks (CNNs). State of the art detectors proposed in the literature follow an end-to-end object detection approach also called single-shot detection, whereby a CNN receives an image as input, scans it only once, and outputs information regarding bounding box and class predictions for the detected objects in the image. One of the most recent approaches, YOLO (You Look Only Once), has shown significant potential for real-time applications compared to approaches based on region proposal \cite{YOLOv2}. Yet the performance drops significantly when moving to embedded platforms \cite{DATELowPowerImage2018}. The YOLO framework works by splitting the input image in a grid and for each grid it generates bounding boxes and class probabilities for a certain number of objects. \textit{DroNet}\cite{DroNet2018} is a network that we have developed that utilizes this framework and is optimized for aerial scene classification for real-time applications with UAVs as shown in Fig. \ref{fig:airYOLO}.

\subsection{Aerial Object Detection using CNNs}
There have been some recent works that attempt to use CNNs for aerial view image object detection. For example, in \cite{RadovicORUAV2017}, a CNN is trained to recognize airplanes in aerial images obtained through digital maps. The framework is based on the YOLO framework and operate on fixed $448\times 448$. The computational cost of the CNN is not addressed and the system is assumed to run on a workstation where the UAV video feed is transmitted to. A CNN for avalanche detection is proposed in \cite{Bejiga.Avalanche.2017} with a Support Vector Machine classifier operating on top of it, and a Hidden-Markov-Model is used as a post-processing step to improve accuracy. However, the target platform is a workstation equipped with a GPU which makes this not suitable for embedding in UAVs. Other works such as \cite{LeeRTUAVCLoud2017} propose cloud computing as a means to achieve real-time object detection from a UAV. The computation of the object detection algorithm is off-loaded to a web service that runs a CNN to detect different objects. This approach is however, primarily intended for short bursts of intense computation and for indoor environments with reliable infrastructure. Also, moving the computation to the cloud introduces communication latencies as well as unpredictable lag which can hinder the application performance. On the other hand, optimizing the approach for on-board processing provides many benefits and can facilitate both indoor and outdoor applications. Other works such as \cite{MariaDroneMELECON2016}, use a sliding-window-based detector  and use a fixed region of interest to discard false positive and reduce the necessary computations. However, this still does not solve the problem of having to scale the detection process to larger resolution images and sliding-window-based detectors are slower that single-shot ones.

It has been shown in the literature that the resolution of the object with respect to the UAV height can affect the detection performance \cite{PetridesICUAS2017}. However, in most works the input image from the UAV camera is first resized prior to performing the detection. Considering the resolutions that UAV cameras operate nowadays, this can lead to problems especially in cases where the UAV flies/hovers at high altitudes where the resolution of the objects of interest becomes small \cite{PetridesICUAS2017}. Hence, it is necessary to develop an appropriate strategy to take advantage of the higher resolution images provided by the UAV camera in order to make object detection more effective but at the same time maintain overall performance at similar levels to that achievable by single-shot detectors for on-board processing.

\section{Proposed Approach}\label{approach}
Reducing the number and size of filters as well as the size of the input image to improve inference speed can lead to considerable loss of information that negatively affects the accuracy. Hence, in this work we explore a new way to boost the detection accuracy of computationally-efficient but resolution-limited CNNs for operating on larger images than their input allows, without changing the underlying network structure. The main idea behind the approach is to separate the larger input image into smaller images with size equal to that of the CNN called \textit{tiles} and selectively process only a subset of them using an \textit{attention} mechanism and track activity in other tiles using a \textit{memory} mechanism. In this way there is no need to resize the input image and the object resolution is maintained.

\subsection{CNN Training and Architecture}
The initial stage of our approach is the collection of training and test data. Images were collected using manually annotated video footage from a UAV and the UCF Aerial Action Data Set \cite{UCFAerialActionDataSet} in order to train two existing CNNs, one based on our previous work \cite{DroNet2018}, and one based on YOLO \cite{YOLOv2}, to detect pedestrians in a variety of scenarios, and different conditions with regards to illumination, viewpoint, occlusion, and backgrounds. The former network is designed to be efficient and lightweight with the main objective of accelerating the execution of the model with minimal compromise on the achieved accuracy. The latter network is a variant of the original YOLO network which was designed for improved inference speeds with some compromise on the accuracy. Overall, for the training set a total of $600$ images were collected with a total of $2000$ pedestrians captured. Each CNN that we tested was trained on the Titan Xp GPU for $200000$ iterations on the same dataset.

\subsection{Selective Tile Processing (STP)}
\begin{figure}[t]
    \centering
    \subfigure{
		\includegraphics[width=0.3\linewidth]{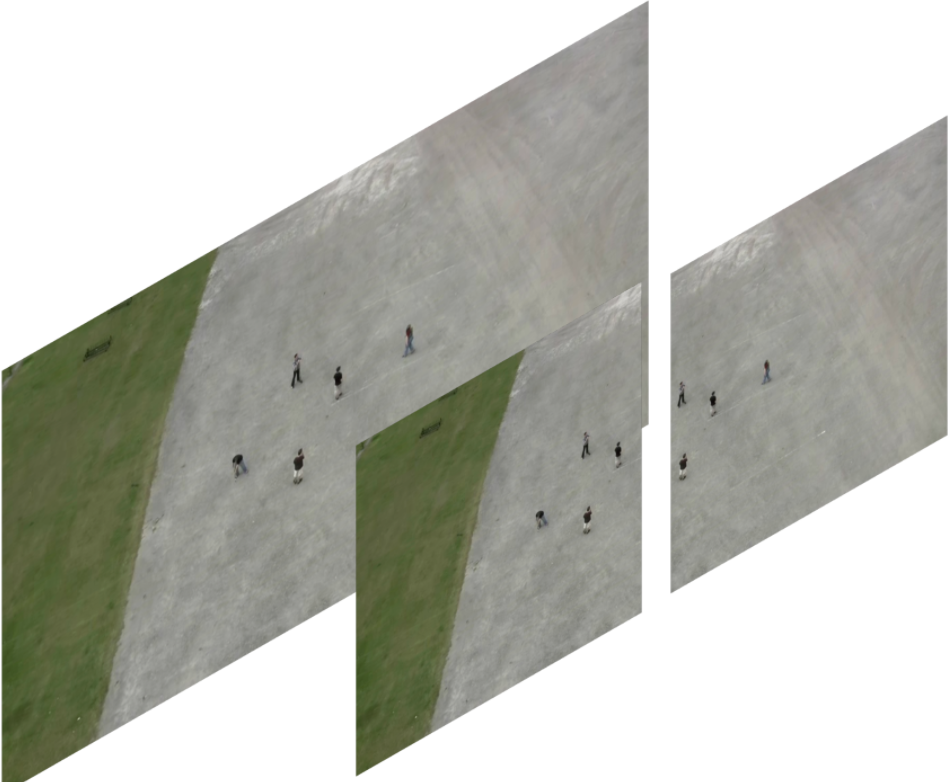}
        \label{fig:544}
       
	}
	\subfigure{
		\includegraphics[width=0.3\linewidth]{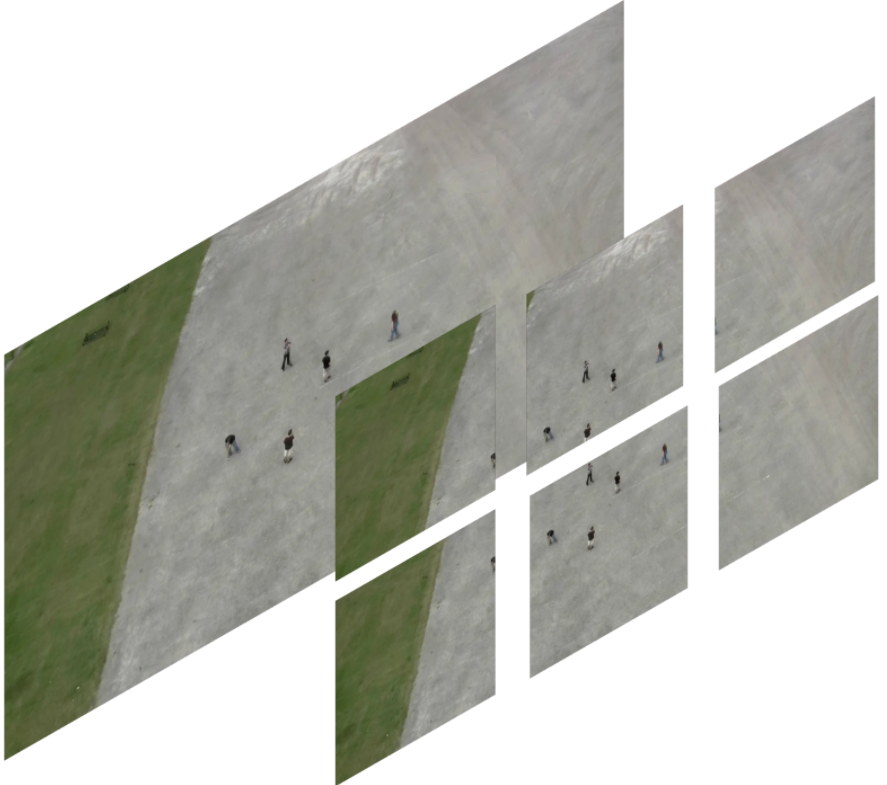}
		\label{fig:352}
		
	}
	\subfigure{
		\includegraphics[width=0.3\linewidth]{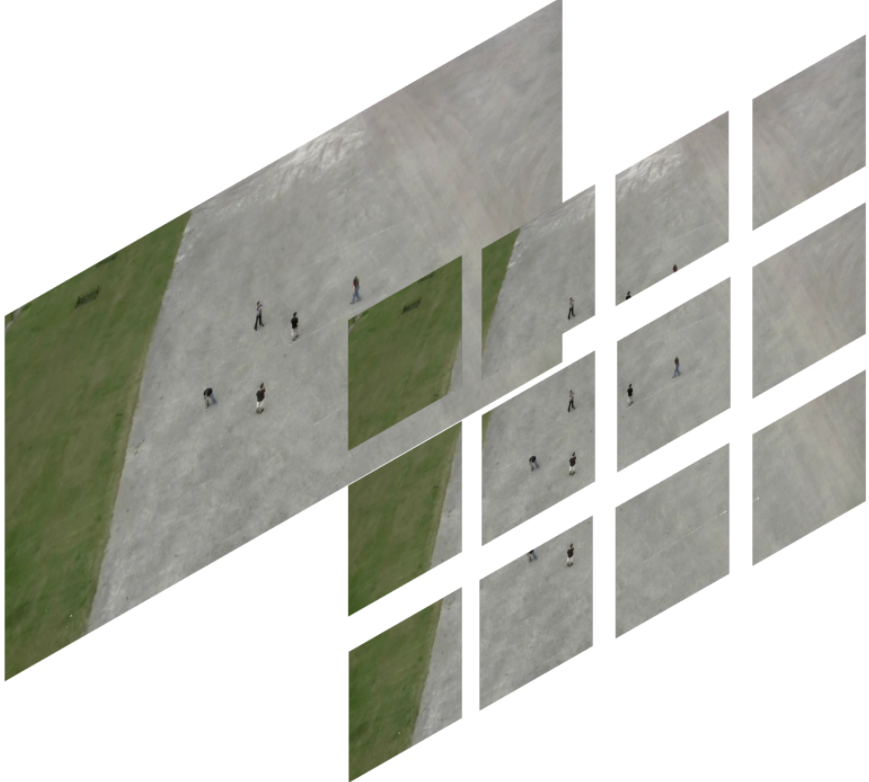}
        \label{fig:256}
        
	}
	\\
	(a)\hspace{0.3\linewidth}(b)\hspace{0.3\linewidth}(c)
	\caption{Tiling for different CNN input sizes: (a) $544\times544$ (2 tiles) (b) $352\times352$ (6 tiles) (c) $256\times256$ (12 tiles)}
	\label{fig:tiles}
\end{figure}

The proposed approach is based on separating the input image into smaller regions capable of being fed to the CNN in order to avoid resizing the input image and maintain object resolution. First, for a given input image size we need to calculate the number of tiles that can be generated. This is done by finding the width and height ratio of the size of the input image with respect to the size of the neural network input (\ref{ratio}). Next these two values are rounded up and multiplied together giving us the number of tiles. After calculating the number of tiles in the horizontal and vertical direction we then uniformly distribute them across the input image so that we achieve complete coverage of the image while maintaining a constant overlap between the tiles as shown in Fig.\ref{fig:tiles}.

\begin{equation}
    \label{ratio}
    Ratio = \frac{Input_{size}}{CNN_{size}}
\end{equation}

Following the calculation of the number of tiles per frame we select which tiles should be processed by the CNN by devising a selective tile processing (STP) strategy based on an \textit{attention} mechanism that utilizes statistical information gathered over time in order to process only a few tiles per frame, and a \textit{memory} mechanism that keeps track of the activity in non-processed tiles.

\subsubsection{Memory Mechanism}
We introduce a memory mechanism  with the objective of using prior information to keep track of objects in tiles that have not be selected for processing in an efficient manner. Instead of using more traditional tracking algorithms that may incur high computational cost and need to tune different parameters and models, we devise a simple yet effective mechanism to remember the position of previously detected targets. A memory buffer is introduced that keeps track of detection metrics in each tile for a number of previous time steps. For each bounding box detected within a tile, useful information is selected that must be stored in order to use this mechanism. This information consists of the position of the bounding box with respect to the image, a detection counter for each bounding box, the latest tile that it was detected in, and the class type (car, pedestrian etc.) if relevant. Only the information for bounding boxes with high confidence are retained and are stored into an appropriate data structure per tile. 

Whenever a bounding box with high confidence is established it must first be categorized as new or belonging to an already detected object. The Intersection over Union (IOU) metric, which captures the similarity between the saved position and the predicted position of the bounding boxes, is utilized to perform this check. Hence, each time a tile is processed by the CNN a comparison using IoU is performed between the currently detected bounding boxes and all the stored boxes while a detection counter keeps track of the times that each box was detected in the past frames. High IoU implies that an already stored bounding box is detected again resulting to the increase of the detection counter for that box and an update of the position of the box. Low IOU implies that the proposed bounding box is a new object appeared in the frame and is stored in the data structure of the detected tile. A stored bounding box which has no overlapping detections over a certain number of past frames is removed.

In this way we have an estimate of the position of the objects in a frame without having to re-process the specific tile that contains the object. We keep the length of our memory buffer relatively small but it can be arbitrarily large depending on the application constrains. By keeping a three frame memory-buffer with the latest positions of the detected objects in the previous frames we can process even a single tile in each frame and retain the position of the objects on the other tiles. The assumption for using this simple but yet effective memory-mechanism is that the relative position of an object is not going to change significantly from the time that we take to process the same tile again. This of course depends on the time needed to process a single tile which depends on the underlying CNN detector. For instance, with a single tile performance of $40$ FPS and $4$ tiles in total each tile will be processed on average with $10$ FPS, thus there is a window $0.1$ seconds before a tile is processed again which is enough for the assumption to hold.

\subsubsection{Attention Mechanism}

The attention mechanism is responsible for selecting which tile(s) to be processed by the CNN. There are different approaches that can be taken in order to carefully choose which tiles to process on the next frame which we outline next:

\textbf{$\textbf{All-Tiles (TA)}:$} The most naive approach is to process all the tiles that will result in improving the accuracy but lead to a large performance degradation on the average processing time. Nevertheless, it provides an indication on the worst possible processing time and acts as a baseline for how much improvement we can gain on the accuracy.

\textbf{$\textbf{Single Tile (T1)}:$} The simplest form of attention is to give every tile the same processing time by switching between tiles in a round robin fashion. It is reasonable to expect that the change in content of successive video frames will be minimal. Therefore, it is highly likely for an object to appear in consecutive frames in more or less the same position as in the previous frame. Thus in the extreme case it might even be sufficient to process a single tile. This approach uses prior information from the memory mechanism to determine the position of each object on the non-processed tiles. Of course this method is agnostic to the activity in each tile while some image regions may need to processed more frequently than others.

The previous two approaches fall between the two extremes of processing all tiles and processing only one tile. On the other hand, it is possible to use the information gathered by processing a tile to guide the overall detection process towards more promising regions. 

\textbf{$\textbf{Select Tiles with Objects (TO)}:$} For this strategy we select only the tiles which contain objects and discard all others. To do that, initially all the tiles are processed so that the number of objects is gathered for all of them. The attention mechanism then discards the tiles with no objects and processes only the tiles where objects have appeared. An issue that arises from this approach is that if nothing is detected in one tile then it will not be processed again. For this reason a \textit{reset time} is introduced after which all the tiles that have not been searched must be fed into the CNN to find objects that may have appeared in the frame during the period that the detector was not looking at those tiles. Of course the value of the \textit{reset time} depends on the movement of the camera and also the movement of the objects in the frame and as such it is determined empirically.

\textbf{$\textbf{Tile Selection and Memory (TSM)}:$} A more intelligent approach is to use the stored information from the memory mechanism and steer the attention mechanism to select the top \textit{N} tiles for further processing instead of selecting a few tiles based only on the number of detected objects in each time step. There are \textit{four} main criteria to assess the value of tile $i$, the number of objects detected in each tile denoted as  $O_{i}$, the cumulative Intersection over Union (IoU), denoted as $I_{i}$, between the processed tile's bounding boxes and the previously saved bounding boxes of an object, the number of times not selected for processing over time denoted as $S_{i}$, and the number of frames past since last detection denoted as $F_{i}$. IoU indicates the movement of the pedestrian with respect to its last detected position with high IoU indicating low movement and low IoU indicating high movement. A selection counter counts the times that each tiles has not been selected for processing and how many frames have gone by without the tile being processed. Moreover, in order to maintain the overall frame-rate above a certain threshold, depending on the application demands, it is efficient to calculated the average processing time for one tile with specific CNN input size and then select as many tiles necessary to maintain that frame-rate. These values are calculated for each tile individually and then normalized by diving with the maximum for each category, effectively reducing their range between $[0..1]$ so that all the different values can be combined together. 

We combine the aforementioned information to select one or more tiles for processing by selecting the ones with the maximum score. The formula in Eq. \ref{selectionmetric} calculates the value of a tile based on the number of detected objects, the cumulative IoU between boxes from the current detection cycle and previously detected boxes, and the selection counter. This score is formulated in such a way so that a high value indicates a promising tile or a tile that has not been recently selected for processing.

\begin{equation}
\begin{gathered}
V_{i} = \dfrac{O_{i}}{max_{j}(O_{j})}+(1-\dfrac{I_{i}}{max_{j}(I_{j})}) + \dfrac{S_{i}}{max_{j}(S_{j})} + \dfrac{F_{i}}{max_{j}(F_{j})}, \\
\forall j\in[0,\dots,(N_{T}-1)] \label{selectionmetric}
\end{gathered}
\end{equation}

\section{Evaluation and Experimental Results}\label{evaluation}
In this section, we present a comprehensive quantitative evaluation of six configurations and strategies as outlined in Section \ref{approach} (using notation $<CNN>^{Method}$):
\begin{itemize}
    \item \textit{$Tiny-YoloV2$}: Resize the input image to the size of the CNN and perform detection only once using the $Tiny-YoloV2$ CNN.
    \item\textit{$DroNet$}: Resize the input image to the size of the CNN and perform detection only once using the $DroNet$ CNN. This small CNN network is also used in the latter approaches.  
    \item \textit{$DroNet^{\textbf{TA}}$}: Process all tiles before moving to the next frame.
    \item \textit{$DroNet^{\textbf{T1}}$}: Process only one tile per frame before moving to the next frame in a Round Robin fashion, and apply memory mechanism to retain information of tiles.
    \item \textit{$DroNet^{\textbf{TO}}$}: Select tiles where object has been detected and use a reset timer.
    \item \textit{$DroNet^{\textbf{TSM}}$}: Use both memory and attention mechanism in order to retain information and select only few tiles to process per frame based on Eq.\ref{selectionmetric}.
\end{itemize}

The basic network models are trained and tested on the same dataset for various input sizes and compared on a CPU on the constructed pedestrian test dataset, consisting of $197$ sequential images extracted from videos in \cite{UCFAerialActionDataSet} containing  $1181$ pedestrians. This dataset allows us to evaluate the performance of the proposed approaches on higher resolution images ($960\times 544$) compared to other approaches in the literature where the image is just resized.

\subsection{Metrics}
To evaluate the performance of each approach, we employ the following two performance metrics:

\textbf{Sensitivity (SEN):}
This is an accuracy metric that gives the percentage of correctly classified objects. It is defined as the proportion of true positives that are correctly identified by the detector and is calculated by taking into account the True Positives ($T_{pos}$) and False Negatives ($F_{neg}$) of the detected objects as given by (\ref{sensitivity}).

\begin{equation}
    \label{sensitivity}
    SEN = \frac{T_{pos}}{T_{pos}+F_{neg}}
\end{equation}

\textbf{Average Processing Time (APT):}
Another important performance metric is the time needed to process a single image frame from a video, which is inversely proportional to the frame-rate or frames-per-second (FPS).
Specifically, this metric is the average processing time across all $N_{test\_samples}$  test images, where $t_i$ is the processing time for image $i$. 
\begin{equation}
    \label{average}
    APT = \dfrac{1}{N_{test\_samples}}\times\sum_{i=1}^{N_{test\_samples}}t_i
\end{equation}

\begin{figure}[t]
\centering
\includegraphics[width=1\linewidth]{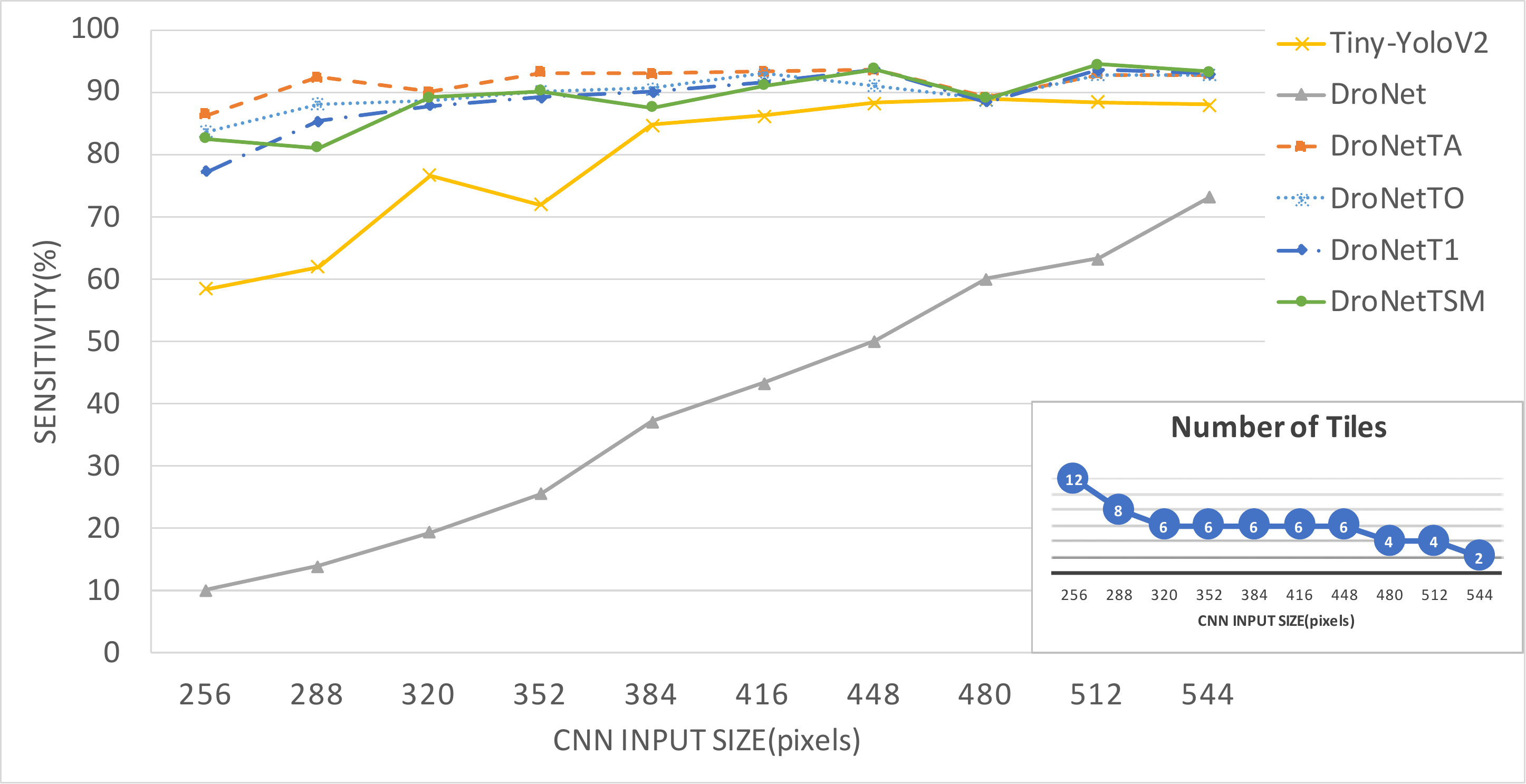}
\caption{Sensitivity for different configurations on CPU platform.}
\label{fig:sens}
\end{figure}

\begin{figure}[t]
\centering
\includegraphics[width=1\linewidth]{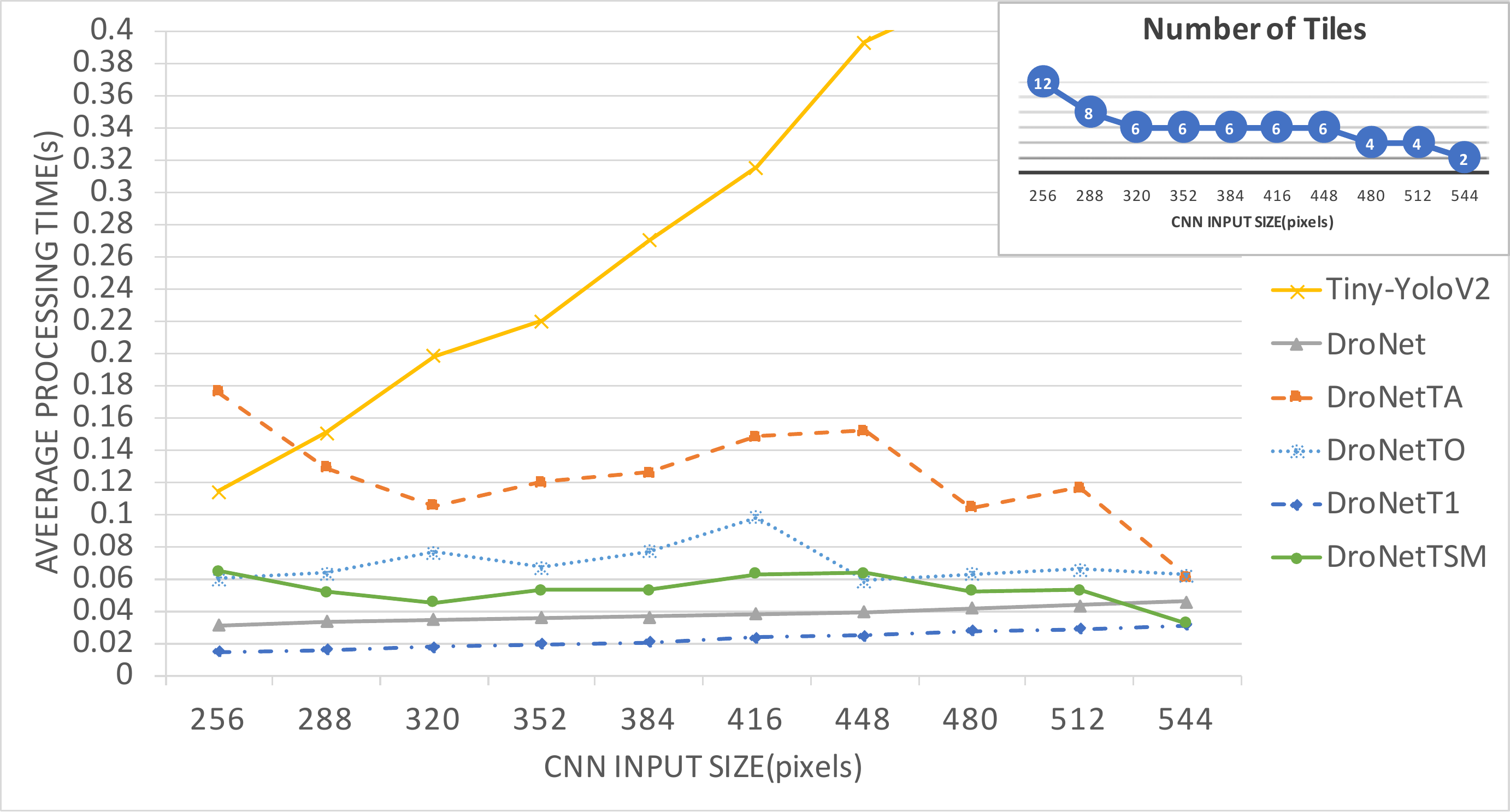}
\caption{}
\label{fig:time}
\end{figure}

\subsection{Performance Analysis on CPU Platform}
We analyzed the impact of the different processing schemes on sensitivity and average processing time. We use a laptop platform with an i5-8250U CPU and 8GB ram for evaluation that can be easily ported to a computational platform used in UAVs such as an Android platform or embedded devices such as Odroid XU4. Fig. {\ref{fig:sens}} and Fig. {\ref{fig:time}} show the sensitivity and the average processing time on the CPU platform with different CNN input image sizes for all approaches.
 
\subsubsection{Baseline approaches}
In our test set with $1181$ pedestrians, \textit{$Tiny-YoloV2$} performs really well with regards to the sensitivity, with maximum $88\%$ and minimum $58\%$. Especially for larger inputs (from $416$ -  $544$) where the object resolution is relatively high. On the other hand, it performs poorly with regards to APT which lies between $0.11 - 0.57 \textit{sec}$ per frame. On the same test set as shown in  Fig. {\ref{fig:sens}}, \textit{$DroNet$} performs $~15\times$ faster compared to \textit{$Tiny-YoloV2$} however, the sensitivity has dropped around $15\%$ as shown in Fig. {\ref{fig:sens}}. It is noticeable that in order to obtain very high sensitivity, large input images must be used which hinders the processing time. For this reason, it is clearly beneficial to employ tiling strategies in order to both improve the sensitivity by processing higher resolution images, and alleviate any negative impact on the average processing time by using a smaller network. 

\subsubsection{\textit{$DroNet^{\textbf{TA}}$} - Processing of all Tiles}
We performed experiments for different number of tiles which varied between $2-12$ with respect to the input size. This tiling method achieved higher sensitivity than both \textit{$Tiny-YoloV2$} and\textit{$DroNet$}, with around $86\%-93\%$ for all CNN input sizes. Fig. {\ref{fig:sens}} and Fig. {\ref{fig:time}} show that by avoiding the resizing of the input image, there is a big improvement on the sensitivity but simultaneously the average processing time was increased by $5 \times$ for all the input sizes. Of course this is to be expected since we increase the workload for the CNN. 

\subsubsection{\textit{$DroNet^{\textbf{T1}}$} - Single tile processing}
In this approach we switch between tiles in a round robin fashion, thus giving equal processing time to all the tiles. The object information is used by the memory mechanism to find the position of the pedestrians on the tiles that were not selected for processing during the past few frames. The results show that the impact on the sensitivity with the memory mechanism is minimal for a small number of tiles($2 - 6$) compared to the previous approaches. Sensitivity is maintained above $75\%$ and is actually as good as processing all tiles for each frame, showing that the memory mechanism, with no impact on performance, can estimate efficiently the position of the pedestrians regardless off CNN input size. On the other hand, the average processing time is tremendously improved compared to the previous approach. There is on average a $2\times$ improvement on speed compared to \textit{$DroNet$}. This can be attributed to the fact that the resizing of a larger image for smaller CNN inputs takes some processing time as well which is not required in our case. Of course, the smaller the size of the input, the higher the number of tiles which has an effect of a larger processing cycle, and can lead to false predictions due to the outdated information in the memory mechanism. Specifically, we have observed an increase of $2-3\times$ in false detections using this approach. Hence, a more elaborate selection process is needed.

\subsubsection{\textit{$DroNet^{\textbf{TO}}$} - Process tiles based on number of objects}
This approach, with a \textit{reset time} of $10$, managed to keep sensitivity above $88\%$ for all input sizes, emphasizing the fact that there is no need to process all the tiles in each frame. On average the processing time was $2 \times$ faster than the previous strategy of processing all tiles for all input sizes; indicating that the number of selected processing tiles in each frame can be reduced even further through the use of more elaborate selection criteria.

\subsubsection{\textit{$DroNet^{\textbf{TSM}}$} - Process tiles based on selection criteria using memory}
The last approach, uses $TSM$ outlined in Section \ref{approach} in order to select a small number of tiles for processing. On average the number of selected tiles is below $35\%$ of the total amount for every CNN input size in order to maintain a steady $APT$ on the CPU platform. There is an increase of $2-3 \times$ in $APT$ compared to  \textit{$DroNet^{T1}$} approach but at the same time a decrease of $2-5 \times$ compared to  \textit{$DroNet^{TA}$}. Moreover, compared to the original \textit{$DroNet$} we managed to maintain similar $APT$ and simultaneously increase the sensitivity by $20-70\%$ across all input sizes. Overall, with $TSM$ we managed to find a good balance between the extreme \textit{$DroNet^{T1}$} approach and the conservative \textit{$DroNet^{TA}$} approach. 

\begin{figure}[t]
\centering
\includegraphics[width=1\linewidth]{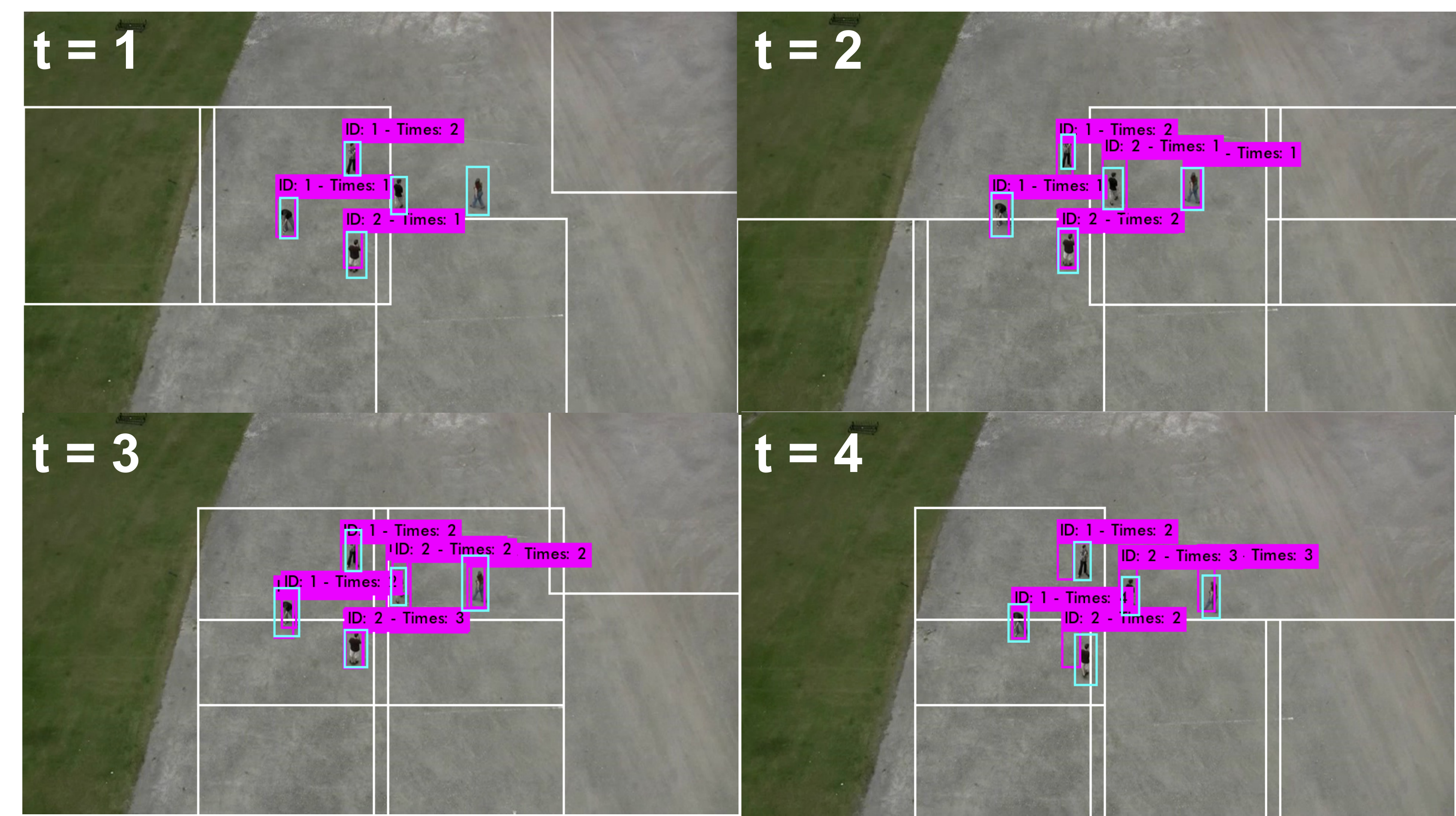}
\caption{A sequence of frames at successive time steps with the tiles that have been selected for processing using the TSM approach. $12$ tiles in total on average $40\%$ selected.}
\label{fig:detection}
\end{figure}

\begin{figure}[t]
\centering
\includegraphics[width=1\linewidth]{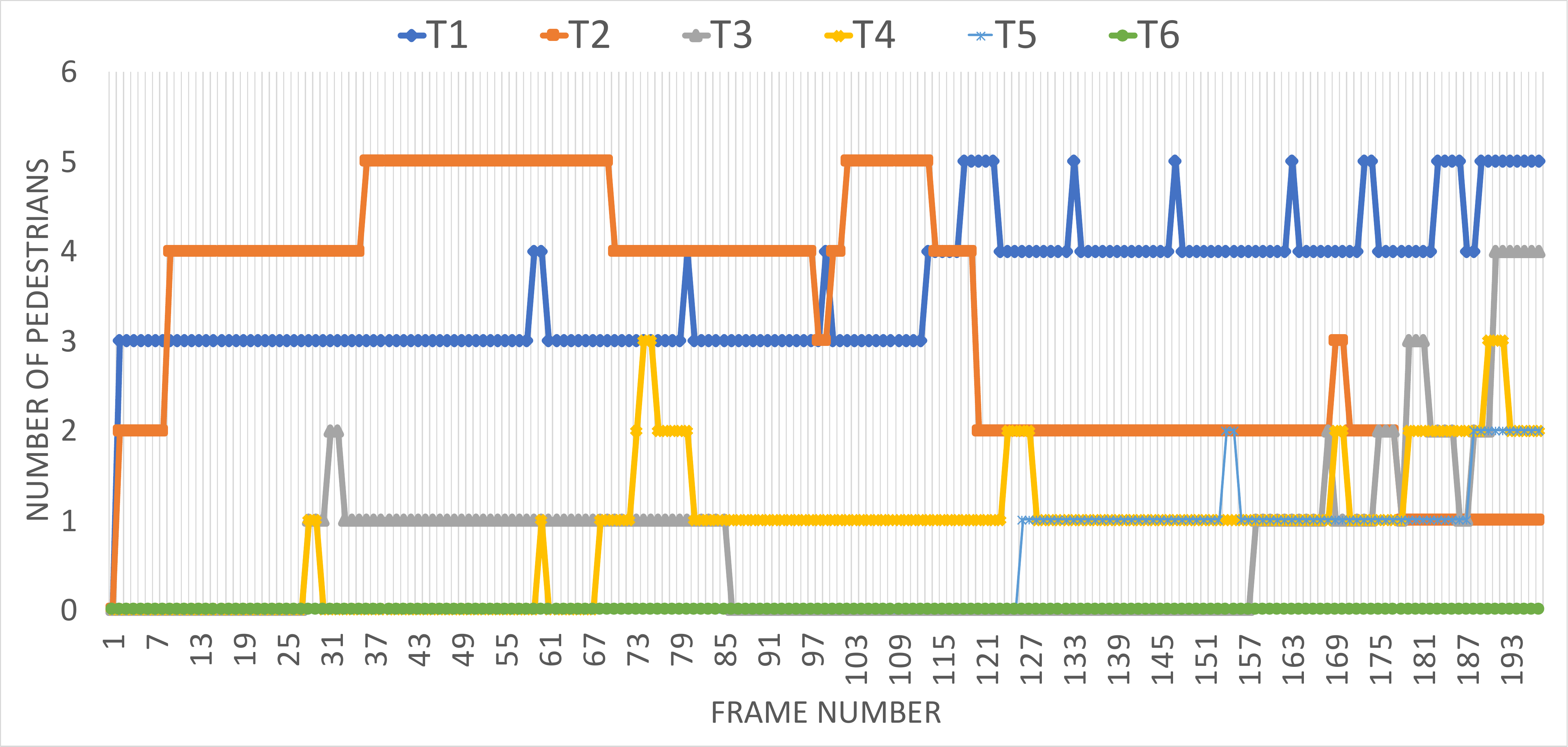}
\caption{Number of Pedestrians for each Tile on every frame.}
\label{fig:ped}
\end{figure}

\begin{figure}[t]
\centering
\includegraphics[width=1\linewidth]{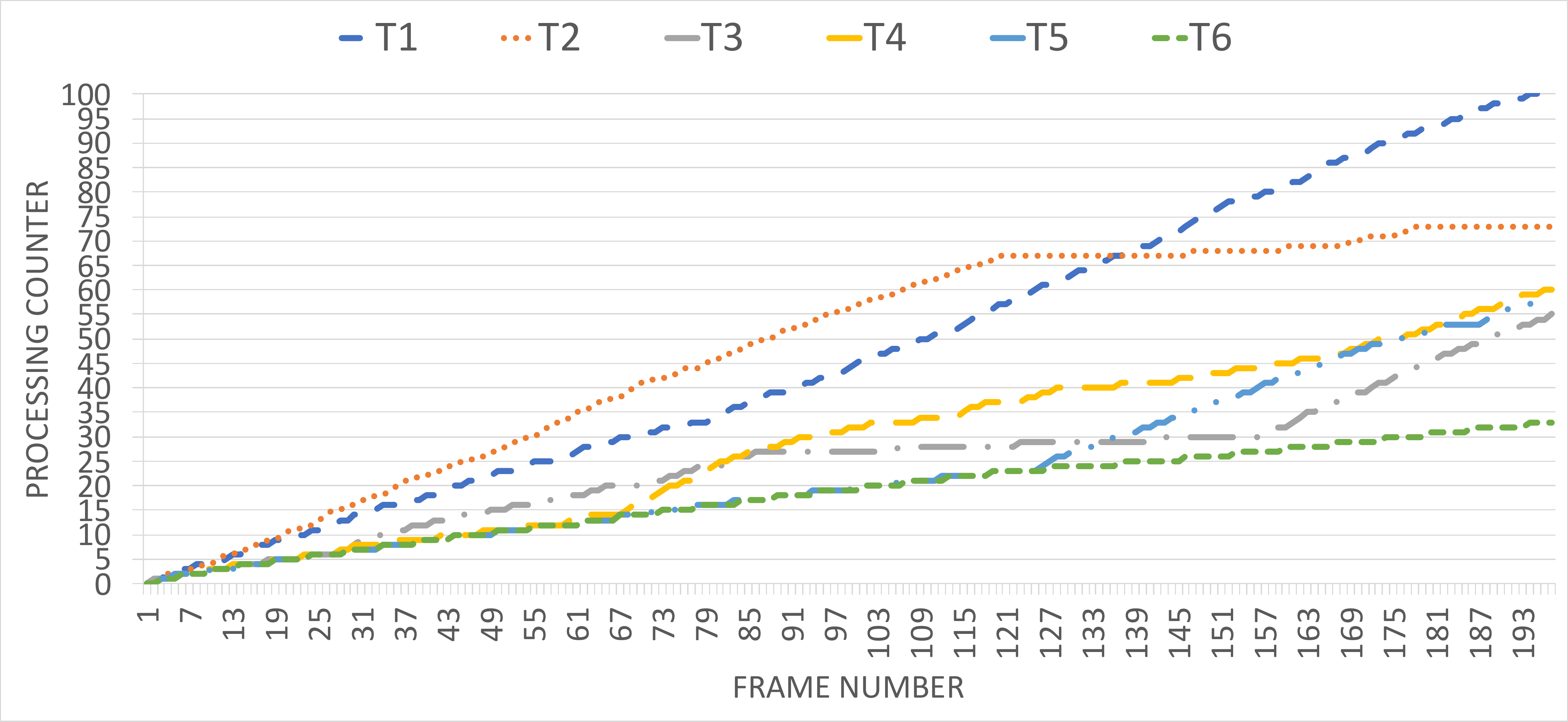}
\caption{Summation of the times that a tile was selected for processing.}
\label{fig:times}
\end{figure}

In addition, we also analyze the $TSM$ approach in terms to which tiles it focuses on and how the attention mechanism selects the tiles during the testing on the dataset. A snapshot of four consecutive frames is depicted in Fig. \ref{fig:detection}, which shows how the attention mechanism selects different tiles for processing based on Eq. \ref{selectionmetric}. Moreover, the information for each tile is extracted and results are presented in Fig.\ref{fig:ped} and Fig.\ref{fig:times}. Fig.\ref{fig:ped} shows the number of detected pedestrians per tile at each time step of the image sequence.  Notice, that overall the number of pedestrians is relatively constant for periods of time due to the effect of the memory mechanism retaining information even if no pedestrian is detected. On the other hand, some transient detections can appear where the memory mechanism takes some frames to discard/accept a detected pedestrian.  Moreover, Fig.\ref{fig:times} shows how many times a tile has been selected for processing. Notice, that the selected tiles (e.g., $T1$) correspond to those that have a high number of pedestrians in Fig.\ref{fig:ped}. Also it is evident that it does not starve those without pedestrians from attention (e.g., $T6$). 


\section{Conclusion \& Future Work}\label{conc}
In this paper we have tackled the problem of efficiently processing higher resolution images using CNNs for real-time UAV smart camera applications. Specifically, we presented a methodology for focusing the detection only on promising image regions without significant degradation in the processing time compared to existing CNN models while improving the accuracy. Through an attention and memory mechanism the proposed approach provides an adequate trade-off by improving accuracy by up to $70\%$ achieving $20$ frames-per-second in a CPU platform. As an immediate follow-up to this work we plan on implementing the $TSM$ approach on computational platforms used in UAVs for further evaluation. Additional optimization's include further tuning parameters and selection criteria to achieve even higher performance and dynamically adjusting the amount, the size, and positioning of the tiles based on the recorded activity. Overall, the approach is general and is suitable for resource constraint systems since it manages to discard large amounts of data and can thus be applied to variety of embedded vision systems such as aerial vehicles to enhance the detection performance on higher resolution images.


\begin{acks}

Christos Kyrkou would like to acknowledge the support of NVIDIA Corporation with the donation of the Titan Xp GPU used for this research.
\end{acks}

\bibliographystyle{ACM-Reference-Format}
\bibliography{references.bib}

\end{document}